\documentclass[12pt]{article}

\usepackage{xcolor}
\usepackage{amssymb}
\usepackage{authblk}

\usepackage{wrapfig}
\usepackage{times}
\usepackage{graphicx}     
\usepackage{amsmath}
\usepackage{txfonts}
\usepackage{bbm}

\usepackage{caption}
\usepackage{subcaption}

\newtheorem{THEOREM}{Theorem}
\newenvironment{theorem}{\begin{THEOREM} \hspace{-.85em} {\bf :} }%
                        {\end{THEOREM}}

\newtheorem{LEMMA}[THEOREM]{Lemma}
\newenvironment{lemma}{\begin{LEMMA} \hspace{-.85em} {\bf :} }%
                      {\end{LEMMA}}
\newtheorem{COROLLARY}[THEOREM]{Corollary}
                          {\end{COROLLARY}}
\newtheorem{PROPOSITION}[THEOREM]{Proposition}
\newenvironment{proposition}{\begin{PROPOSITION} \hspace{-.85em} {\bf :} }%
                            {\end{PROPOSITION}}
\newtheorem{DEFINITION}[THEOREM]{Definition}
\newenvironment{definition}{\begin{DEFINITION} \hspace{-.85em} {\bf :} \rm}%
                            {\end{DEFINITION}}
\newtheorem{CLAIM}[THEOREM]{Claim}
                            {\end{CLAIM}}
\newtheorem{EXAMPLE}[THEOREM]{Example}
                            {\end{EXAMPLE}}
\newtheorem{REMARK}[THEOREM]{Remark}
                            {\end{REMARK}}
							\newtheorem{NOTATION}[THEOREM]{Notation}
							                            {\end{NOTATION}}
\newenvironment{proof}{\noindent {\bf Proof:} \hspace{.677em}}%
                     {}

\newcommand{\bbox}{\vrule height7pt width4pt depth1pt}
\newcommand{\qed}{\bbox\vspace{0.1in}}
\DeclareMathAlphabet{\mathitbf}{OML}{cmm}{b}{it}

\newcommand{\real}{{\mathbb{R}}}

\newcommand{\blemma}{\begin{lemma}}

\newcommand{\elemma}{\end{lemma}}

\newcommand{\bthm}{\begin{theorem}}
\newcommand{\ethm}{\end{theorem}}
\newcommand{\bprf}{\begin{proof}}
\newcommand{\eprf}{\end{proof}}
\newcommand{\bpro}{\begin{proposition}}
\newcommand{\epro}{\end{proposition}}
\newcommand{\bi}{\begin{itemize}}
\newcommand{\ei}{\end{itemize}}
\newcommand{\be}{\begin{enumerate}}
\newcommand{\ee}{\end{enumerate}}
\newcommand{\beq}{\begin{equation}}
\newcommand{\eeq}{\end{equation}}
\newcommand{\bcase}{\begin{cases}}
\newcommand{\ecase}{\end{cases}}
\renewcommand{\mathit}{\emph}

\usepackage{ragged2e}
\usepackage{mathtools}

\usepackage[utf8]{inputenc} 
\usepackage[T1]{fontenc}    
\usepackage{hyperref}       
\usepackage{url}            
\usepackage{booktabs}       
\usepackage{amsfonts}       
\usepackage{nicefrac}       
\usepackage{microtype}      
\usepackage{xcolor}         

\title{Principled Diverse Counterfactuals in Multilinear Models}
\date{\vspace{-5ex}}
\author[1]{Ioannis Papantonis\thanks{i.papantonis@sms.ed.ac.uk}}
\author[1,2]{Vaishak Belle\thanks{vbelle@ed.ac.uk}}

\affil[1]{The University of Edinburgh}
\affil[2]{Alan Turing Institute}

\begin{document}

\maketitle

\begin{abstract}
Machine learning (ML) applications have automated numerous real-life tasks, improving both private and public life. However, the black-box nature of many state-of-the-art models poses the challenge of model verification; how can one be sure that the algorithm bases its decisions on the proper criteria, or that it does not discriminate against certain minority groups? In this paper we propose a way to generate diverse counterfactual explanations from multilinear models, a broad class which includes Random Forests, as well as Bayesian Networks. \end{abstract}

\section{Introduction}
In recent years explanations for machine learning (ML) models have gained a lot of prominence, especially in the context of safety critical applications. This is due to the black-box nature of many of the state-of-the-art models, which impedes a thorough understanding of their internal reasoning. One of the most important challenges is how can one be sure that the algorithm bases its decisions on the proper criteria, or that it does not discriminate against certain minority groups? This kind of questions have motivated the development of research fields such as explainable AI (XAI) \cite{DBLP:journals/corr/abs-2009-11698} and fairness in AI \cite{10.1145/3457607}.

XAI aims at addressing these concerns by developing techniques and appropriate measures that allow for examining a model's intrinsic behaviour. 
A specific approach that has received considerable attention is to use counterfactual explanations, a minimal set of modifications that are sufficient to make a model change its decision. This line of research has led to the development of a general framework for producing counterfactual instances without resorting to approximations, described in Wachter et al \cite{Wachter2017CounterfactualEW}. Building on top of that, \cite{10.1145/3351095.3372850} proposes a method for generating diverse counterfactuals for differentiable models. Nevertheless, subsequent works, such as \cite{10.1145/3287560.3287569} address some technical challenges, proposing a new framework that is based on mixed integer programming (MIP) in order to remedy the stability issues of the aforementioned approaches.

Additional benefits of utilizing the methodology in \cite{10.1145/3287560.3287569} is that it provides for a principled way to generate diverse counterfactuals. However, a major limitation is that it is only applicable to linear models, thus its range of applications is quite narrow.
Our work focuses on extending this methodology to a broad class of non-linear architectures, namely (ensembles of) multilinear models, which include decision trees (DTs), random forests (RFs), and Bayesian network classifiers (BNCs). This way we both expand the scope of the existing method, as well as provide a way for generating counterfactuals for non-differentiable models, something that is not possible using \cite{10.1145/3351095.3372850,Wachter2017CounterfactualEW}. We show how this problem can be expressed as an integer linear program (ILP) that is guaranteed to generate valid counterfactuals. Furthermore, we draw connections with other existing approaches and, in fact, show that some of them correspond to a special case of our proposed framework. Moreover, when it comes to DTs and RFs, our method results into a (possibly infinite) set of counterfactuals. This is in tune with other recent works that generate counterfactuals for these models \cite{ijcai2018-708}, however we arrive at this point from a different perspective. Finally, we discuss how one can seamlessly generate diverse counterfactuals using our framework.



\section{Related work}

Counterfactuals have a long standing history within philosophy \cite{Lewis1974-LEWC-20,Ruben1990-RUBEE-3}, as well as within the causal modelling community \cite{10.5555/1642718}. When it comes to XAI, they have gained significant traction in recent years, partly because there is evidence suggesting that non-technical audience feels more comfortable interpreting such explanations over alternatives, such as propositional rules \cite{10.1145/3173574.3173951}. Furthermore, counterfactuals inherently convey a notion of ``closeness'' to the actual world, in the sense that they allow for detecting a set of minimal changes that can alter a model's decision. In the seminal work of Wachter et al \cite{Wachter2017CounterfactualEW}
 an optimization scheme for generating counterfactuals is proposed, based on Lagrange multipliers, assuming the classifier is differentiable. 

On the other hand, Russel \cite{10.1145/3287560.3287569} proposes a different framework, based on MIP, to generate counterfactuals for linear models. As the author notes, this resolves the technical issues of \cite{Wachter2017CounterfactualEW}, while it provides a principled way for generating diverse counterfactuals, too, since utilizing only a single counterfactual can be overly restrictive \cite{Wachter2017CounterfactualEW}, impeding a better understanding of the model.


Apart from the aforementioned approaches, the problem of generating counterfactuals instances has been considered from alternative angles as well. In a recent line of work \cite{ijcai2018-708,shi2020tractable,DBLP:journals/corr/abs-2007-01493}, a different framework for producing counterfactuals, among others, is presented, based on utilizing tractable architectures, such as OBDDs \cite{10.1145/136035.136043}. At the core of these works lies the idea of transforming a classifier into another structure, which allows for answering a number of queries in polynomial time. However, we should note that the transformed model can be exponentially larger than the original one. Regardless of that, this framework provides for a principled way for generating counterfactuals for some non-differentiable models which cannot be handled utilizing the approaches in \cite{Wachter2017CounterfactualEW,10.1145/3351095.3372850}.

\section{Background}

In this section we are going to briefly introduce the models we are going to utilize in the following.


\begin{figure}
\centering
\begin{minipage}{.3\textwidth}
  \centering
  \includegraphics[scale=1]{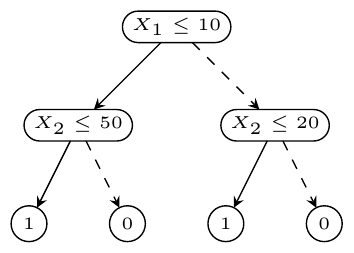}
  \captionof{figure}{A Decision Tree}
  \label{fig:test1}
\end{minipage}%
\begin{minipage}{.7\textwidth}
  \centering
  \includegraphics[scale=0.33]{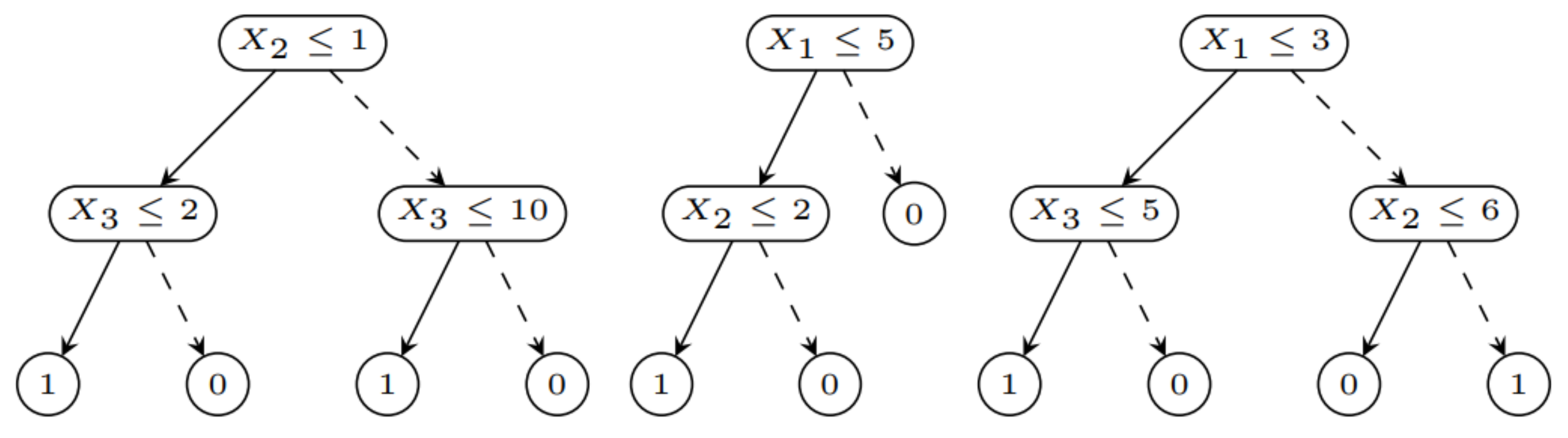}
  \captionof{figure}{A Random Forest}
  \label{fig:test2}
\end{minipage}
\end{figure}

\subsection{Decision Trees}

Decision trees (DTs)  are tree-like structures that contain a  set  of  conditional  control  statements, such as $X\leq a$. Each assignment is consistent with exactly one root-to-leaf path, corresponding to the model's outcome. The control statements are arranged  in  a  hierarchical  manner,  where intermediate nodes represent decisions and leaf nodes can be either class labels (for classification problems) or continuous quantities (for regression problems).

The majority of decision tree learning algorithms operate in a top-down manner, iteratively partitioning the whole dataset into smaller ones, conditioning on the values of the feature that contains the most information, in each iteration. This has led to the development of a number of metrics that quantify the amount of information that is gained, when splitting the dataset according to a specific feature, such as Gini impurity \cite{10.5555/1162264} In turn, these metrics can be used in order to design algorithms that learn DTs from data, such as CART \cite{https://doi.org/10.1002/cyto.990080516}.

An advantage of employing DTs is that their internal rule-based architecture is relatively easy to inspect, allowing for assessing the quality of the model. This is one of the major reasons why DTs are usually utilized in cases where the model's understandability is essential, or in fields like medicine. However, large DTs containing a lot of rules are not easy to interpret anymore, requiring additional explainability tools in order to reason about their internal behaviour \cite{DBLP:journals/corr/abs-2009-11698}.

\subsection{Random Forests}

As we discussed in the previous section, DTs have been employed in various applications due to the transparency they exhibit, at least as long as they are kept at a reasonable size. However, one of their major limitations is their tendency to overfit the given dataset, leading to high variance models that fail to maintain good performance when dealing with new data. 

Random forests (RFs) aim at overcoming this challenge by combining multiple trees, resulting in more stable models with lower variance. The main insight underlying this approach is to sample with replacement from the whole dataset in order to construct multiple new datasets, thus implementing the idea of bagging \cite{10.1023/A:1018054314350}. Following that, a decision tree is trained over each of these newly acquired datasets, leading to an ensemble of independent trees. Then, in prediction time, an aggregation measure, such as majority voting (for classification) or averaging (for regression), combines the predictions of each tree in order to generate the prediction of the whole forest.

The procedure described above results in very expressive and accurate models, however this comes at the expense of interpretability, since the whole forest is far more challenging to explain, compared to single decision trees. This has led to the development of various techniques that attempt to explain the inner reasoning of a RF \cite{DBLP:journals/corr/abs-2009-11698}.


\begin{figure}
\centering
\begin{minipage}{.45\textwidth}
  \centering
  \includegraphics[scale=1]{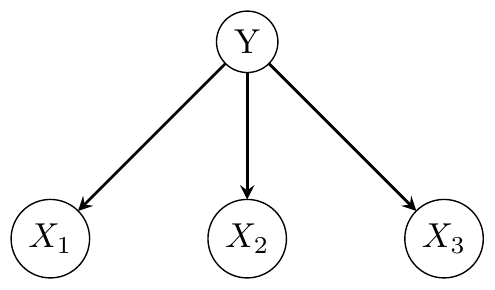}
  \captionof{figure}{A Naive Bayes model as BN.}
  \label{bn}
\end{minipage}%
\begin{minipage}{.55\textwidth}
  \centering
  \includegraphics[scale=0.5]{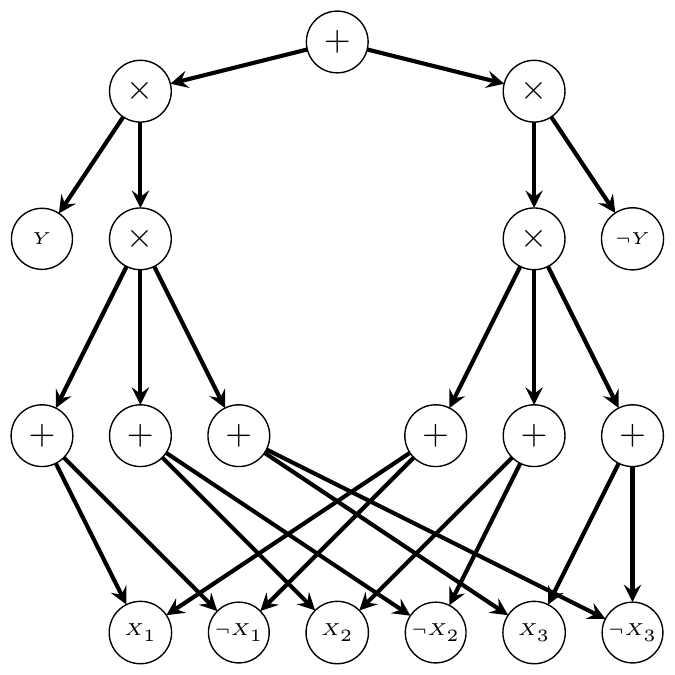}
  \captionof{figure}{A naive Bayes model as SPN.}
  \label{spn}
\end{minipage}
\end{figure}

\subsection{Sum-Product Networks}
Sum-product networks (SPNs) are rooted directed graphical models that provide for an efficient way of expressing a joint distribution that is defined over a Bayesian network (BN). Assuming all variables are binary (or categorical, in general) SPNs encode this distribution as a multilinear function, $\sum_{\textbf{x}}f(\textbf{x})\prod_{n=1}^{N} \mathbbm{1}_{x_n}$. Here $f(\cdot)$ is the  (possibly unormalized) probability distribution of the BN, $\textbf{x}$ is a vector containing all the variables of the model, i.e.,  $x_1,\cdots,x_N$, the summation is over all possible states, and $\mathbbm{1}_{x_n}$ is the indicator function \cite{10.1145/765568.765570}. In its simplest form, this function contains $2^N$ terms, however, when context-specific independence among the variables is present, it is possible to obtain a compact factorized representation, that is not exponential in the number of the model's variables.

SPNs are strictly more efficient than distributions that are defined over BNs using CPDs, since any such distribution can be transformed to a SPN in polynomial time and space, while the converse is not true \cite{pmlr-v37-zhaoc15}. Furthermore, SPNs generalize a number of well known models \cite{10.5555/3044805.3044886}, such as thin junction trees \cite{10.5555/2980539.2980614} and latent tree models \cite{10.5555/1953048.2021056}. On top of that, computing marginal or conditional probabilities in SPNs is linear in its size, making them an appealing candidate for practical applications. Since we are considering classification problems, we are interested in discriminative SPNs \cite{10.5555/2999325.2999496}, that encode the conditional distribution of a target variable given some predictors, while they also subsume Bayesian network classifiers (BNCs) \cite{10.1145/765568.765570}.For example, figures (\ref{bn}, \ref{spn}) show the two different representations of a naive Bayes classifier.



\section{Problem Derivation}
In this section we introduce our approach for generating counterfactuals, inspired by \cite{10.1145/3287560.3287569}, but addressing one of its key limitations; the range of models it applies to. Specifically, we extend the existing framework to multilinear models, such as DTs and BNCs, as well as ensembles thereof that utilize majority voting, such as RFs. In what follows we assume that all variables are binary, to allow for an easier presentation. However, we provide an extension to the non-binary case, in section \ref{non}.



Before going any further, we begin with defining a quantity similar to the decision function, developed in \cite{ijcai2018-708}, as follows:
\begin{definition}
Let $G:\textbf{X}\rightarrow\{0,1\}$ be a binary classification function, and $P^G_0(\textbf{X}),P^G_1(\textbf{X})$ be multilinear polynomials of indicator variables, where all coefficient are equal to $1$ and there is no constant term.  Then $P^G_0(\textbf{X})$ (respectively, $P^G_1(\textbf{X})$) is called the 0-decision (resp. 1-decision) polynomial of $G$, iff $G(\textbf{X})=0\Leftrightarrow P^G_0(\textbf{X})=1~~(\text{resp. }G(\textbf{X})=1\Leftrightarrow P^G_1(\textbf{X})=1)$
\end{definition}

Decision polynomials provide for a multilinear representation of arbitrary binary classifiers. In \cite{ijcai2018-708}, decision functions play a similar role, however there is no requirement for them to be multilinear. In our work, we have this additional condition in order to be able to derive an optimization problem in ILP format. 

In the remaining of this section, we derive some results that hold for decision polynomials, in general. In the following subsections we make the necessary adjustments to apply the developed framework to DTs, RTs and BNCs. The next proposition follows immediately from the definition, and will be used extensively throughout the rest:
\begin{proposition}\label{prop}
Let $G:\textbf{X}\rightarrow\{0,1\}$, and $P^G_0(\textbf{X}),P^G_1(\textbf{X})$ be the decision polynomials. Then $\forall \textbf{x} \in \textbf{X}~~P^G_0(\textbf{x})+P^G_1(\textbf{x})=1$
\end{proposition}



The following statement is a simple observation that since each term of a decision polynomial is equal to either 0 or 1, in order for the polynomial to output 0, each term has to be equal to 0.

\begin{proposition}\label{lem}
Let $G:\textbf{X}\rightarrow\{0,1\}$, and $P^G_0(\textbf{X}),P^G_1(\textbf{X})$ be the decision polynomials. Assuming $P^G_0(\textbf{X})=T_1(\textbf{X})+T_2(\textbf{X})+\dots + T_k(\textbf{X})$, where each $T_i \in \{0,1\}$, then $P^G_0(\textbf{X})=0\Rightarrow T_1(\textbf{X})=T_2(\textbf{X})=\dots = T_k(\textbf{X})=0$. The same holds for $P^G_1(\textbf{X})$.
\end{proposition}

Proposition \ref{lem} implies that in order to make sure that a decision polynomial outputs $0$, it is enough to make sure that each monomial equals $0$. The next challenge is due to the fact that these monomials are products of indicator functions, not linear combinations of them. This situation impedes the formulation of generating counterfactuals as a linear optimization problem. A key insight for overcoming this difficulty is that since indicator functions can be equal to either $0$ or $1$, making sure that not all of them are equal to $1$ is sufficient to guarantee that their product is equal to $0$. The following proposition states a simple condition that leads to this outcome.

\begin{proposition} \label{notall}
Let $X_1,X_2,\cdots \cdot X_k \in \{0,1\} $, then $X_1\cdot X_2 \cdots X_{k-1} \cdot X_k =1 \Rightarrow X_1 + X_2 + \cdots + X_k = k $ and $X_1\cdot X_2 \cdots X_{k-1} \cdot X_k =0 \Rightarrow X_1 + X_2 + \cdots + X_k \leq k-1 $.
\end{proposition}

At this point, propositions \ref{lem} and \ref{notall} already provide for a set of constraints that are sufficient to ensure that a datapoint is classified as either $0$ or $1$. For example, if the goal is to generate an instance that belongs in the $1$-class, then it is enough to consider the $0$-decision polynomial and for each term, say $X_1\cdot X_2 \cdots X_{k-1} \cdot X_k$, add the constraint $X_1 + X_2 + \cdots + X_k \leq k-1$. This procedure guarantees that the solution to the problem, $\textbf{X}$, satisfies $P^G_0(\textbf{X})=0 \Rightarrow P^G_1(\textbf{X})=1$, so it is classified as $1$. 

However, having said that, storing both polynomials requires additional resources, while it could also be the case that one of them is significantly smaller than the other one, so it would be preferable to express the problem in terms of this polynomial to end up with a more compact optimization problem. A natural way to address this situation would be to define a set of constraints that when satisfied force a term in the decision polynomial to be equal to $1$, and the rest equal to $0$. The following proposition provides such a set of constraints:

\begin{proposition}\label{proposition}
Let $P_0(\textbf{X})=X_{11}\cdot X_{12}\cdots X_{1k} + X_{21}\cdot X_{22}\cdots X_{2m} + \cdots + X_{n1}\cdot X_{n2}\cdots X_{nl}$, where each $X_i \in \{0,1\}$, be the $0$-decision polynomial of a model. Furthermore, let the constraints $X_{11} + X_{12}\cdots + X_{1k}  \geq k \cdot \delta_1, X_{21} + X_{22}\cdots + X_{2m} \geq m \cdot \delta_2, \dots, X_{n1} + X_{n2}\cdots + X_{nl}  \geq l\cdot \delta_n, \sum_{i=1}^n \delta_i = 1 ,$  where $\delta_i \in \{0,1\}$. If an assignment, $\textbf{X'}$, satisfies these constraints, then $P_0(\textbf{X'})=1$. An analogous statement holds for $P_1(\textbf{X})$.
\end{proposition}

We have now developed most of the the necessary machinery to formulate a counterfactual generating optimization problem. In summary, these are:
\begin{itemize}
    \item Construct one of the two decision polynomials. Let us assume we construct the $0$-DP.
    \item Form the objective function.
    \item If the counterfactual instance has to be classified as $0$, apply proposition \ref{notall} to every term of the DP.
    \item If the counterfactual instance has to be classified as $1$, utilize the constraints in proposition \ref{proposition} to enforce this outcome.
\end{itemize}

In the following subsections we address the first two points, providing ways to recover the DPs of DTs, RFs and BNCs, as well as discussing various ways to define an optimization function. Furthermore we provide some adjustments that need to be made in order to take into account the characteristics of the aforementioned models.

\subsection{Decision trees}\label{dt}

Decision trees can be naturally seen as a collection of rules, so in this section we will examine how this set of rules can be used in order to construct a tree's decision polynomial. Transforming DTs to equivalent rule-based classifiers is a well studied problem \cite{10.1016/S0020-7373(87)80053-6}. They key observation however, is that it is possible to derive a multilinear representation of a DT over the set of rules it naturally induces.



An example of the general process can be seen in figure (\ref{fig:test1}), which contains a very simple decision tree. It is defined over two continuous variables, $X_1,X_2$, but it can also be seen as a function over its internal rules, $X_1\leq 10, X_2\leq 50, X_2\leq 20$. Utilizing the latter, and traversing the DT bottom-up, it is not difficult to see that the decision polynomials are:
\begin{align*}
    & P^G_1(X_1,X_2)= \mathbbm{1}[X_1\leq 10]\cdot\mathbbm{1}[X_2\leq 50] + (1-\mathbbm{1}[X_1\leq 10])\cdot\mathbbm{1}[X_2\leq 20],\\
    & P^G_0(X_1,X_2)=\mathbbm{1}[X_1\leq 10]\cdot(1-\mathbbm{1}[X_2\leq 50]) + (1-\mathbbm{1}[X_1\leq 10])\cdot(1-\mathbbm{1}[X_2\leq 20]),
\end{align*}
where $\mathbbm{1}$ is the indicator function. 

The $1$-DT contains all the rules that the DT utilizes to classify an instance in the $1$-category, while the $0$-DT follows an analogous reasoning. In both polynomials, all monomials are monic, as well as there is no constant term. Furthermore, since for each possible assignment only one root-to-leaf path will be satisfied, each polynomial outputs either $0$ or $1$, so they are indeed valid DPs. This process exemplifies the general reasoning, which remains unaltered, no matter how large a DT is. 

Having the decision polynomials, we are now ready to put all the pieces together. To this end, let $d=(X_1,X_2)$ be a factual datapoint of interest. We utilize the weighted $l_1$ norm, $\Vert\cdot\Vert_{1,w}$, and the rule representation of the DT to define the distance between two points as follows:
 \begin{align*}
 \Vert d-d' \Vert_{1,w}&=w_1|\mathbbm{1}[X_1\leq 10]-\mathbbm{1}[X_1'\leq 10]| +w_2|\mathbbm{1}[X_2\leq 20]-\mathbbm{1}[X_2'\leq 20]| \\
 &+w_3|\mathbbm{1}[X_2\leq 50]-\mathbbm{1}[X_2'\leq 50]|,
 \end{align*}
where $w_1,w_2,w_3$ are constants. This is the objective function of the final optimization problem.
The last step is to remove the absolute values from the objective function. This is simple to do, since the values of the indicators  $\mathbbm{1}[X_1\leq 10], \mathbbm{1}[X_2\leq 20], \mathbbm{1}[X_2\leq 50]$ are known quantities, and $0\leq \mathbbm{1}[\cdot]\leq 1$.

To go on with our example let us also assume that $d(X_1,X_2)$ satisfies $X_1\leq 10, 20 <X_2 \leq 50$, so it is classified  into the $1$ class, and that we want to utilize the $0$-DP. Applying proposition \ref{proposition}, the final optimization problem is:
\begin{align*}
&min~~ w_1(1-\mathbbm{1}[X_1'\leq 10]) + w_2\mathbbm{1}[X_2'\leq 20] +w_3(1-\mathbbm{1}[X_3'\leq 50])~~~s.t.\\
&\mathbbm{1}[X_1'\leq 10] + (1-\mathbbm{1}[X_2'\leq 50]) \geq 2\cdot\delta_1, ~~~
(1-\mathbbm{1}[X_1'\leq 10)] + (1-\mathbbm{1}[X_2'\leq 20]) \geq 2\cdot \delta_2,\\
&\delta_1+\delta_2=1
\end{align*}

The solution of this problem is guaranteed to be classified as $0$. Of course it depends on the values of $w_1,w_2,w_3$, but it is going to be an infinite set of solutions, regardless. For example, if the resulting solution turns out to be $\mathbbm{1}[X_1'\leq 10]=1, \mathbbm{1}[X_2'\leq 20]=\mathbbm{1}[X_2'\leq 50]=0$, then every element of the set $\{(X_1,X_2'):X_2'> 50 \}$ is a valid counterfactual to $d$, with respect to the decision tree. This is an extension of the framework in \cite{10.1145/3287560.3287569}, where the outcome was a single point.



Finally, we discuss the amount of constraints that has to be added within the model. As the DPs encode the root-to-leaf paths of the decision tree, the amount of constraints depends on the number of distinct root-to-leaf paths, $m$. The added flexibility of expressing our framework using either of the two DPs, allows for efficiently handling situations that would be otherwise problematic. For example, if there is a DT having only one path that leads to a $0$-leaf, and all the remaining ones lead to a $1$-leaf, then we can encode everything using the $0$-DP in a highly efficient manner, using a single constraint, instead of $m-1$ ones. This demonstrates that the worst-case scenario is when there is an equal number of $0$-leaf and $1$-leaf paths, in which case the cost of encoding the constraints is the same, no matter which DP is utilized. This means that in the worst case $O(\frac{m}{2})$ constraints would be necessary, each one involving $O(p)$ variables, where p is the length of the longest path in the tree.


\subsection{Random Forests}


In this section, we examine how to handle ensembles of multilinear models, using RFs as an example. Although the process is similar in spirit, incorporating information from multiple models poses an additional challenge. For example, looking at figure (1b) we can  verify that the 1-DP of each tree is:
\begin{align*}
T_1: &P_1^{T_1}(X_1,X_2,X_3)= \mathbbm{1}[X_2\leq 1]\cdot \mathbbm{1}[X_3\leq 2] + (1-\mathbbm{1}[X_2\leq 1]) \cdot \mathbbm{1}[X_3\leq 10],\\
T_2: &P_1^{T_2}(X_1,X_2,X_3)= \mathbbm{1}[X_1\leq 5]\cdot \mathbbm{1}[X_2\leq 2],\\
T_3: &P_1^{T_3}(X_1,X_2,X_3)= \mathbbm{1}[X_1\leq 3]\cdot \mathbbm{1}[X_3\leq 5] + (1-\mathbbm{1}[X_1\leq 3]) \cdot (1-\mathbbm{1}[X_2\leq 6])
\end{align*}

As usual, each individual polynomial encodes all the $0$ or $1$ assignments of each individual tree, but how can we combine them all together so they encode the behaviour of the forest? A first remark is that in order to make sure that the model outputs, for example, $1$, it is enough to enforce a constraint that at most one $0$-decision polynomial outputs $1$. This would mean that the outcome of at least $2$ out of the $3$ decision trees is equal to $1$, so the whole forest has an output of $1$, assuming majority voting.

This kind of reasoning can be applied to ensembles will an arbitrary number of trees, and will be the base of extending the current framework. As a matter of fact, it turns out that this approach corresponds to a generalization of the one we presented for DTs (see supplementary material for details). The following proposition provides for a way to encode the fact that a decision polynomial is equal to $1$. 




\begin{proposition}\label{coro}
Let $P_0^G(\textbf{X}) = X_{11}\cdot X_{12}\cdots X_{1k} + X_{21}\cdot X_{22}\cdots X_{2m} + \cdots + X_{n1}\cdot X_{n2}\cdots X_{nl}$, where each $X_i \in \{0,1\}$, be the $0$-decision polynomial of a model. Furthermore, let the constraints $X_{11} + X_{12}\cdots + X_{1k} - k \leq \delta -1, X_{21} + X_{22}\cdots + X_{2m} -m \leq \delta -1, \cdots, X_{n1} + X_{n2}\cdots + X_{nl} - l \leq \delta -1$, where $\delta \in \{0,1\}$. Then, $\delta=0 \Rightarrow P_0^G(\textbf{X})=0$. An analogous statement holds for $P_1^G(\textbf{X})$.
\end{proposition}

Proposition \ref{coro} can be used as an indicator of whether a DT outputs $0$ or $1$, but it can be easily extended so it applies to a RF. For example, assuming we utilize the $0$-DPs to generate an instance that is classified as $1$, adding this set of constraints for every DT in the RF and demanding that at least half of the corresponding indicators are equal to $0$, we enforce that the majority of the DTs have an outcome equal to $1$, so the whole forest outputs $1$.

Furthermore, as it was the case with DTs, utilizing proposition \ref{coro} and both DPs it is now possible to state all the necessary constraints to ensure the desired outcome. However, the same considerations as before apply to the RF case, so it would be desirable to be able to express the optimization problem in terms of a single DP. As it turns out, it is possible to extend proposition \ref{proposition} so it can handle the RF case as well:
\begin{proposition}\label{rf}
Let $T_1,T_2,\cdots,T_m$ be the DTs of a RF $F$. For each $T_j$, consider $P_0^{T_j}(\textbf{X})$ and add all the constraints appearing in proposition \ref{proposition}, except for the last one, which is replaced by $\sum_{i=1}^n \delta_{ji} = \delta_j$, where $\delta_{ji}$ appears in the i-th constraint of the j-th tree and $\delta_j \in \{0,1\}$ is a newly introduced variable. Finally, add the constraint $\sum_{i=1}^m \delta_i > \left\lfloor \frac{m-1}{2} \right\rfloor$. If an assignment, $X'$, satisfies these constraints, then $P_0^{F}(X')=1$.
\end{proposition}



These results connect the behaviour of a single model to the behaviour of the ensemble, allowing to control the number of models that output a certain outcome. However, tree ensembles present an additional challenge that needs to be addressed; that is, we need to make sure that the solution of the optimization problem is consistent. In this setting, we use the term consistency in the sense that if the solution dictates that a condition of the form $X\leq \alpha$ holds, then all the conditions of the form $X\leq \beta$, where $\alpha\leq \beta $ hold as well. Furthermore, by the same reasoning, if a condition $X\leq \alpha$ does not hold, then no condition $X\leq \beta$, where $\alpha\geq \beta $ should hold. To this end, we have the following definition:



\begin{definition}
Let $T_1,T_2,\cdots,T_n$ be DTs and $X$ one of the variables in their scope. We define:
\begin{itemize}
    \item $F_x = \{ X\leq a |~ X\leq a \in F(T_i), ~i \in \{1,2,\cdots,n\} \}$, where $F(T_i)$ is the set of all the internal rules in $T_i$. In turn, $F_x$ is the set of all the rules among all the trees that involve variable $X$.
    \item Furthermore, let $X\leq a$ be an element of $F_x$, and define $F_x^+ (X\leq a) = \{ X\leq b|~X\leq b \in F_x, b\geq a\}$, the set of rules involving $X$ where the threshold is larger than $a$, and $~F_x^- (X\leq a) = \{ X\leq b|~X\leq b \in F_x, b\leq a\}$, the rules where the threshold is smaller than $a$.
\end{itemize}

\end{definition}


The following proposition provides a way to achieve consistency by enforcing a set of constraints:
\begin{proposition}\label{prop8}
Let $T_1,T_2,\cdots,T_n$ be DTs that form a RF. Then, the constraints $\sum_{f_i \in F_x^+ (X\leq a)} \mathbbm{1}[f_i] \geq |F_x^+ (X\leq a)| \cdot \mathbbm{1}[X\leq a]$ and $\sum_{f_i \in F_x^- (X\leq a)} \mathbbm{1}[f_i] \leq |F_x^- (X\leq a)|\cdot \mathbbm{1}[X\leq a]$, guarantee that the final solution is consistent wrt the feature $X\leq a$.
\end{proposition}

Looking at proposition \ref{prop8} we see that two constraints per feature are enough to guarantee  consistency. We can now examine the number of constraints that are required in order to generate a counterfactual set from a RF. Clearly, we have to include the counterfactual generating constraints as well as the consistency ones. The former, amounts to incorporating the DP of each tree in the forest. As discussed in the previous section, assuming there are $N$ trees, $O(\frac{Nm}{2})$ constraints are required in the worst case, where $m$ is the maximum number of distinct paths among all $N$ trees. For the latter, we have to add  two constraints per feature,meaning that $O(NF^*)$, where $F^*={max}_{T_1,\cdots,T_N}(F_{T_i})$, constraints are required. Combining these together, in the worst case $O(N(\frac{m}{2}+F^*))$ constraints are needed to define a counterfactual generating problem.

We are now ready to demonstrate how to generate counterfactuals for RFs, by combining proposition \ref{coro}, proposition \ref{proposition}, and proposition \ref{prop8}. Returning to our running example, let $d=(X_1,X_2,X_3)$ be a datapoint that satisfies the conditions $X_1\leq 3,~X_2\leq 1,~X_3 \leq 2$, meaning that all $3$ DTs classify $d$ as $1$. Assuming we utilize the $1$-DPs, the following generates a set of counterfactuals that are classified as 0:
\begin{align*}
 &\min w_1 (1 - \mathbbm{1}[X_1 \leq 5]) + w_2 (1 - \mathbbm{1}[X_1 \leq 3]) + w_3 (1 - \mathbbm{1}[X_2 \leq 1]) + w_4 (1 - \mathbbm{1}[X_2 \leq 2]) \\
&  + w_5 (1 - \mathbbm{1}[X_2 \leq 6]) + w_6 (1 - \mathbbm{1}[X_3 \leq 2]) + w_7 (1 - \mathbbm{1}[X_3 \leq 10]) + w_8 (1 - \mathbbm{1}[X_3 \leq 5])~~~\text{s.t.} \\
&\begin{rcases}
 &\mathbbm{1}[X_2 \leq 1] + \mathbbm{1}[X_3 \leq 2] -2 \leq \delta_1 - 1,\\
 &1 - \mathbbm{1}[X_2 \leq 1] + \mathbbm{1}[X_3 \leq 10] - 2 \leq \delta_1 - 1\\
&\mathbbm{1}[X_1 \leq 5] + \mathbbm{1}[X_2 \leq 2] - 2 \leq \delta_2 - 1,\\
&\mathbbm{1}[X_1 \leq 3] + \mathbbm{1}[X_3 \leq 5] - 2 \leq \delta_3 - 1 \\
& 1 - \mathbbm{1}[X_1 \leq 3] + 1 - \mathbbm{1}[X_2 \leq 6] - 2 \leq \delta_3 - 1, \\
&\delta_1 + \delta_2 + \delta_3 \leq 1\\
\end{rcases}\text{0 generating constraints}\\
&\begin{rcases}
&\mathbbm{1}[X_1 \leq 5] \geq \mathbbm{1}[X_1 \leq 3]\\
\end{rcases}\text{$X_1$ consistency constraints}\\
&\begin{rcases}
&\mathbbm{1}[X_2 \leq 2] + \mathbbm{1}[X_2 \leq 6] \geq 2\mathbbm{1}[X_2 \leq 1], ~~~\mathbbm{1}[X_2 \leq 6] \geq \mathbbm{1}[X_2 \leq 2]\\
&\mathbbm{1}[X_2 \leq 1] \leq \mathbbm{1}[X_2 \leq 2], ~~~\mathbbm{1}[X_2 \leq 1] + \mathbbm{1}[X_2 \leq 2] \leq 2 \mathbbm{1}[X_2 \leq 6]\\
\end{rcases}\text{$X_2$ consistency constraints}\\
&\begin{rcases}
&\mathbbm{1}[X_3 \leq 5] + \mathbbm{1}[X_3 \leq 10] \geq 2\mathbbm{1}[X_3 \leq 2], ~~~\mathbbm{1}[X_3 \leq 10] \geq \mathbbm{1}[X_3 \leq 5]\\
&\mathbbm{1}[X_3 \leq 2] \leq \mathbbm{1}[X_3 \leq 5], ~~~\mathbbm{1}[X_3 \leq 2] + \mathbbm{1}[X_3 \leq 5] \leq 2 \mathbbm{1}[X_3 \leq 10]
\end{rcases}\text{$X_3$ consistency constraints}
\end{align*}

\begin{table}[t]
\centering
\scalebox{0.75}{\begin{tabular}{|l|c|c|c|c|c|c||c|}
\hline
 & Sex & Age & Race & Juvenile felonies &  Prior crimes & Two year residivism & Outcome\\
\hline
Factual & Male & 33 & Caucasian & 0 & 2 & Yes & Low score\\
Counterfactual  & Male & 33 & Caucasian & \textbf{>0} & 2 & Yes & High score\\
Diverse counterfactual & \textbf{Female} & 33 & Caucasian & (\underline{=0}) 0 & 2 & Yes & High score\\
\hline
\end{tabular}}
\caption{A COMPAS dataset instance}
\label{compas}
\end{table}

\subsection{Sum-Product Networks}

The last class of models we are going to incorporate within our framework are BNCs over binary variables, representing them as SPNs. Retrieving the DPs of an SPN is relatively straightforward, utilizing their interpretation as a collection of tree models \cite{NIPS2016_6c9882bb}. Of course, this means that in the worst case an SPN is a collection of an exponential amount of trees, one for each joint variable assignment. In turn, this means that in such a scenario an exponential amount of constraints is needed in order to encode a decision polynomial. This is in tune with known complexity results that utilize tractable structures to compute counterfactuals \cite{ijcai2018-708}.

Despite that, SPNs have been particularly powerful in applications where there is context-specific independence \cite{10.5555/2074284.2074298} among the variables, providing very compact representations. This means that although the worst case scenario is exponential, there are situations where it is possible to define the optimization problem using significantly fewer constraints. For example, let us assume that the 0-DP of the SPN in figure (\ref{spn}) is equal to $P_0(X_1,X_2,X_3)= X_1X_2X_3 + X_1X_2(1-X_3)$. The terms in the polynomial imply that when $X_1=X_2=1$, a datapoint is classified in the 0 class, regardless of what value $X_3$ has. In turn, we end up with the reduced 0-DP  $P_0(X_1,X_2,X_3)= X_1X_2$.


The above process can be repeated iteratively, eliminating variables that are not relevant, given some context, just like $X_3$ was irrelevant, given the context $X_1=X_2=1$. A simple way to achieve this elimination is whenever encountering two terms differing in only one factor, to substitute both of them with a new term that is equal to their common factors. For example, let us assume that this time the 0-DP is equal to $P_0(X_1,X_2,X_3)= X_1X_2X_3 + X_1X_2(1-X_3) + X_1(1-X_2)X_3 + X_1(1-X_2)(1-X_3)$. It is not difficult to observe that this polynomial is equal to $1$, only when $X_1=1$, meaning it can be reduced to a simpler form. Applying our strategy leads to:
\begin{align*}
   P_0(X_1,X_2,X_3)&= \underbrace{X_1X_2X_3 + X_1X_2(1-X_3)}_\text{$X_3$ is irrelevant, given $X_1=X_2=1$} + \underbrace{X_1(1-X_2)X_3 + X_1(1-X_2)(1-X_3)}_\text{$X_3$ is irrelevant, given $X_1=1,X_2=0$}\\
   &= \underbrace{X_1X_2 + X_1(1-X_2)}_\text{$X_2$ is irrelevant, given $X_1=1$} = X_1,
\end{align*}
which exactly matches our observation. In the same way, we can handle decision polynomials, in general, possibly leading to a significant reduction in size, whenever sufficient context-specific information is available.

\subsection{Parameters, the non-binary case and diversity} \label{non}

In this section we will discuss some approaches to set the weights, $\textbf{w}$, in the $\ell_1$ norm as well as some possible extensions. In the original work of Watcher et al \cite{Wachter2017CounterfactualEW}, the inverse of the median absolute deviation (MAD) of a feature is utilized. As the authors argue, some advantages of this particular choice is that it captures the intrinsic volatility of a feature, as well as it is more robust to outliers, compared to using the standard deviation. However, MAD is inappropriate when using binary features, since in this case it is always equal to zero \cite{10.1145/3287560.3287569}. Regardless, in the supplementary material we provide a way to work around this issue for DTs and RFs, while for BNCs we agree that the inverse standard deviation is a sensible choice.



A special case worth mentioning arises when all weights are equal to $1$. Then, the resulting distance, $\Vert \cdot\Vert_{1,\textbf{1}}$, reduces to the Hamming distance \cite{shi2020tractable}, and the solution of the minimization problem reflects the smallest number of changes that are necessary for the model to change its output. As a matter of fact, this number has already gained significant attention within an emergent line of research regarding explainability approaches in Bayesian classifiers, where it is known as the \textit{robustness} of a classifier \cite{shi2020tractable}. However, existing methods are applicable only when utilizing the Hamming distance, which does not allow for assigning different weights to features. In this sense, our framework extends current approaches, since it allows for computing the robustness of a classifier under alternative metrics, that admit non-uniform feature weights, reflecting the relative importance of each term. 

We would also like to note that although we have focused on how to generate counterfactuals, it is possible to generate alternative forms of explanations, by making a few minor adjustments. In the appendix we explain how to compute \textit{prime implicant (PI) explanations} \cite{ijcai2018-708}. Unlike counterfactual explanations that compute a minimal set of changes enough to alter the model's decision, PI-explanations compute a minimal set of feature values that is enough to maintain the model's decision, no matter the values of the remaining features. Furthermore, as it was the case with the counterfactual explanations, the proposed framework allows for assigning non-uniform weights to each feature, something that is not possible using symbolic approaches, such as \cite{shi2020tractable}.

So far we have assumed that all variables (or rules) are binary, but it is possible to extend our framework to the non-binary case, by utilizing a simple transformation. In general, let $X$ be a variable taking values in $\{0,1,\cdots,k\}$. We can now introduce $k$ new binary variables, $X_0,X_1,\cdots, X_k$, such that $X_i=1 \Leftrightarrow X=i$. Furthermore, we need to add the constraint $\sum_{i=0}^k X_i =1$ to enforce that $X$ takes exactly one value. Employing this trick it is immediate to handle the non-binary case.


One of the benefits of our proposal is that it is seamless to generate diverse counterfactuals. This is not true for many of the existing techniques, but it is a benefit of employing ILP, as recognized by Russel \cite{10.1145/3287560.3287569}. In the BNC case it is as simple as just setting the variables to their desired values, leaving everything else intact. In the DT and RF cases, since variables variables may be continuous, a user could ask for counterfactuals that satisfy conditions such as $a\leq X\leq b$, for a variable $X$ (or a set of variables). This is again easy to handle, since the condition $a\leq X\leq b$, is enough to decide the values of some of the constraints in $F_x$. Then, it is just a matter of plugging these values into the optimization problem and proceeding as normal, leaving the rest unchanged. 

\begin{table}[t]
\centering
\scalebox{0.75}{\begin{tabular}{|l|c|c|c|c|c|c||c|}
\hline
 & Sex & LSAT & Race & UGPA  & Outcome\\
\hline
Factual & Male & 34 & White & 3 & Pass\\
Counterfactual  & Male & \textbf{< 19.25} & White & 3 & Fail\\
Diverse counterfactual & Male & (\underline{>25}) 34 & \textbf{Black} & \textbf{<1.95} &  Fail\\
\hline
\end{tabular}}
\caption{A LSAT dataset instance}
\label{lsat}
\end{table}

\section{Experiments}
In this section we will demonstrate some of the advantages of utilizing the proposed framework. To this end, we will examine three different case studies, based on the COMPAS, LSAT, and  Congressional Voting Records datasets. For the first two, we are going to employ a DT and a RF, respectively, while for the last one we use a Naive Bayes Classifier, although any BNC can be used. Due to space limitations, we present only a single case for each dataset, however additional ones can be found in the supplementary material. Counterfactual conditions are in bold, while diversity conditions are inside a parenthesis.

\textbf{COMPAS: }COMPAS is a popular algorithm for assessing the likelihood that a person will reoffend (recidivate) within two years from being released from prison. It has drown significant attention within the fairness in AI community, due to the number of biases it exhibits, such as favoring white inmates against black ones \cite{doi:10.1126/sciadv.aao5580}. The dataset contains the COMPAS training variables (age, race, sex, number of prior crimes, number of juvenile felonies), whether the inmate actually reoffended within a two-year period (two year residivism), as well as the final score generated by the algorithm. A DT was trained on this dataset, predicting the risk of reoffending.

Table \ref{compas} shows the record of an inmate, where the first row represents the factual datapoint, the second an unconstrained counterfactual, and the third one is making use of the diversity constraints. A first remark is that the unconstrained counterfactual is in fact an infinite counterfactual set, since any instance satisfying ``juvenile felonies > 0'' is a valid counterfactual. 

At this point, judging from the unconstrained counterfactual alone, it would be difficult to assess whether the model exhibits any bias. However, this is a case where diversity constraints can lead to valuable insights. The constraint we enforced was ``number of juvenile felonies = 0''. The resulting counterfactual now uncovers the model's biased behavior, since it suggests that had the inmate been female, the model would predict a high score of reoffending.  

\textbf{LSAT: } LSAT is another popular dataset in the fairness literature, since it exhibits a strong bias against black people, too. In this setting, the model has to predict whether students will pass the bar, based on their sex, age, law school admission test (lsat), and undergraduate gpa (ugpa). Table \ref{lsat} shows a student record, along with the model's prediction. The unconstrained counterfactual is again an infinite set, where making sure that lsat is less than 19.25 is enough to alter the model's prediction. 

While looking at this counterfactual does not reveal any biases, the relative discrepancy between the factual value of lsat and the counterfactual condition (about 15 points), should be an indicator that constraining the lsat value closer to its factual value, could expose biased behavior. While incorporating inequality constraints is in general very challenging, in our framework it reduces to assigning specific values to some of the indicator variables. As it turns out, enforcing that lsat is greater than 25 leads to a counterfactual that clearly showcases the bias in the model, since the student's race is a factor that can alter the model's decision.

This case also demonstrates how such counterfactuals can be used to guide an inspection of the dataset in order to identify the reasons behind this behavior. Looking at it, we see that $96.7\%$ of male, white students passed the bar, while the same percentage for male, black students was $77.8\%$. Furthermore, the number of white students in the dataset was about $21$ times bigger than that of black ones. This shows that black, male students are severely under-represented, while the imbalance between successful/unsuccessful students in the two groups may lead the model to assign significant predictive power to a student's race.

Having said that, utilizing the counterfactual it is possible to perform a more targeted analysis, to uncover imbalances that are not as apparent. To this end, we inspected the dataset for black, male students with $lsat < 19.5$ and $gpa = 3$, only to find out that all such students failed to pass the bar. However, for white, male students, with the same characteristics, half of them passed the bar. On top of this discrepancy, even the specific instances prompt biased behavior, since, for example, a black student with $lsat = 19$, $gpa = 3$, failed, while a white one with $lsat = 17.5$, $gpa = 3$, succeeded, encouraging the model to take racial information into account.

Following this analysis, it should come as no surprise that the RF picked up a corresponding bias, since by looking at the individual DTs we found out that there are $6$ different paths that lead to a positive outcome for white, male students with $gpa < 3$, as opposed to only $1$ for black students. This means that the RF is more ``forgiving'' towards white students with low gpa, in contrast to black ones. Targeting these two specific subgroups was guided by the insights we obtained from the counterfactual, which eventually led to the discovery of significant information, regarding both the dataset and the model.

\textbf{ Congressional Voting Records: } This dataset contains the votes of the U.S. House of Representatives Congressmen on 16 key votes. This time, the problem is to predict  whether a person is a Democrat or a Republican, based on these 16 votes. To this end, we trained a Naive Bayes classifier, however the same analysis can be performed for any BNC. Table \ref{votes} shows how a particular congressman voted (where + represents voting for, and - voting against). This time, instead of computing counterfactuals, we will present prime implicants explanations, as generated by the proposed framework. 

The unconditional prime implicants form a set of 4 elements, meaning that as long as the votes regarding topics 3, 4, 5, 14 remain the same, the model will always classify the person as a Democrat.
Furthermore, to further inspect the model, it is possible to compute conditional prime implicants. For example, requiring that the first vote remains the same, we see that the resulting explanation now has 5 elements, some of them not present in the unconditional explanation. This result indicates there is some relationship among these variables, which could in turn motivate additional analysis.

\begin{table}[t]
\centering
\scalebox{0.75}{\begin{tabular}{|l|c|c|c|c|c|c|c|c|c|c|c|c|c|c|c|c||c|}
\hline
 & 1 & 2 & 3 & 4 & 5 & 6 & 7 & 8 & 9 & 10 & 11 & 12 & 13 & 14 & 15 & 16 & Outcome\\
\hline
Factual & + & + & + & - & - & - & + & + & + & - & + & - & - & - & + & + & Democrat\\
Prime implicants  &  &  & \checkmark & \checkmark & \checkmark &  &  &  &  &  &  &  &  & \checkmark &  &  & Democrat\\
Conditional prime implicants  & \checkmark  &  &  & \checkmark & \checkmark &  &  &  &  &  & \checkmark  &  &  & \checkmark &  &  & Democrat\\
\hline
\end{tabular}}
\caption{A Congressional Voting Records dataset instance}
\label{votes}
\end{table}

\section{Future work and conclusions}

In this work we present a framework for generating counterfactual explanations for (ensembles of) multilinear models. This way we extend the methodology in \cite{10.1145/3287560.3287569}, as well as generalize some of the results in \cite{ijcai2018-708}. We show how to apply our results to DTs, RFs, and BNCs, but any multilinear model can be utilized, instead. This is in contrast to methods like, \cite{FERNANDEZ2020196}, since this is based on a modification of the CART algorithm, so it is only applicable to DTs and RFs. Analogously, for BNCs, we show how our framework permits more expressive distance functions, that incorporate the relative importance of each term, instead of treating all feature changes as equally important or feasible.


In our opinion there are a lot of interesting research directions to go from here. A first remark is that as can be seen from the complexity results, the worst case scenario is exponential, so there are cases where encoding a DP can be impractical. These situations highlight the importance of developing approximate representations of DPs, that correctly classify instances with high probability. This seems like a natural next step, especially considering the long-standing research line of approximate reasoning in BNs, as well as some recent attempts at approximate reasoning with DTs and RFs \cite{intrees}. Other interesting directions include defining probabilistic versions of DPs, reflecting how probable an assignment is, since currently all assignments are treated as equally probable. Advances in these areas could facilitate generating out-of-the-box counterfactuals, leading to their wider use in practical applications.


\bibliographystyle{abbrv}
\bibliography{arxiv}

\clearpage
\section{Appendix}
\subsection{RF constraints generalize DT constraints}
We start by showing that the optimization schema for RTs is indeed a generalization of the DT one. Without loss of generality, we can assume that we are using the $0$-DP to formulate the optimization problem, since the same argument applies to the other case as well. Let $T$ be a DT, and let us first consider the case of generating an instance that is classified as $1$. To this end, we will utilize the constraints in proposition 6, treating $T$ as a trivial RF, $F$, comprised of just a single tree. 

The $0$-DP of $F$ is identical to the $0$-DP of $T$, so  $P_0^{F}(X)=P_0^{T}(X) = X_{11}\cdot X_{12}\cdots X_{1k} + X_{21}\cdot X_{22}\cdots X_{2m} + \cdots + X_{n1}\cdot X_{n2}\cdots X_{nl}$. Following the procedure in proposition 6, it suffices to add the constraints:
\begin{align*}
    &X_{11} + X_{12}\cdots + X_{1k} - k \leq \delta_1 -1\\
    &X_{21} + X_{22}\cdots + X_{2m} -m \leq \delta_1 -1\\
    &~~~~~~~~~~~~~~~~~~\cdots\\
    &X_{n1} + X_{n2}\cdots + X_{nl} - l \leq \delta_1 -1\\
    &\delta_1\leq 0
\end{align*}

Of course, since $\delta_1\geq 0$, the constraint $\delta_1\leq 0$ implies that $\delta_1=0$. Substituting this into the remaining constraints leads to an updated set of constraints:
\begin{align*}
    &X_{11} + X_{12}\cdots + X_{1k}  \leq k -1\\
    &X_{21} + X_{22}\cdots + X_{2m}  \leq m -1\\
    &~~~~~~~~~~~~~~~~~~\cdots\\
    &X_{n1} + X_{n2}\cdots + X_{nl}  \leq l -1\\
\end{align*}
These are exactly the constraints that result from propositions 3 and 4, thus establishing the desired equivalence for this case.

Furthermore, we have to examine the case of generating an instance that is classified as $0$, using the $0$-DT. Again, treating $T$ as a trivial RF, we utilize proposition 7 this time to obtain the set of sufficient constraints:
\begin{align*}
    &X_{11} + X_{12}\cdots + X_{1k}  \geq k\cdot\delta_1\\
    &X_{21} + X_{22}\cdots + X_{2m} \geq m\cdot\delta_2\\
    &~~~~~~~~~~~~~~~~~~\cdots\\
    &X_{n1} + X_{n2}\cdots + X_{nl} \geq l\cdot\delta_n\\
    &\sum_{i=1}^n\delta_i = \delta\\
    &\delta>0
\end{align*}
The last constraint implies that $\delta=1$, so we can rewrite the constraints as:
\begin{align*}
    &X_{11} + X_{12}\cdots + X_{1k}  \geq k\cdot\delta_1\\
    &X_{21} + X_{22}\cdots + X_{2m} \geq m\cdot\delta_2\\
    &~~~~~~~~~~~~~~~~~~\cdots\\
    &X_{n1} + X_{n2}\cdots + X_{nl} \geq l\cdot\delta_n\\
    &\sum_{i=1}^n\delta_i = 1
\end{align*}
Again, these are exactly the constraints in proposition 5. Additionally, since there is only a single tree in forest, it is not necessary to include the consistency constraints, because inconsistencies only arise when combining multiple trees. This concludes the proof of the claim that the RF constraints generalize the DT ones.

\subsection{MAD weights for DTs and RFs}
The MAD of a feature is defined as:
\begin{equation}\label{mad}
\text{MAD}_k = \text{median}_{j \in D}(|X_{j,k}- \text{median}_{l \in D}(X_{l,k})|)
 \end{equation}
where $D$ is the dataset, and $X_{i,k}$ denotes the value of feature $k$, in data point $i$.

While MAD is a sensible choice when using continuous data, it is inappropriate when using binary features, since in this case it is always equal to zero. However, for the DT and RF cases, although the optimization problem is expressed in terms of binary variables, reflecting the nature of their intrinsic splitting rules, the variables themselves can be continuous. This results in an interesting situation, where both the MAD and the standard deviation are valid weighting options. For example, consider a feature $k$ and branching rule of the form $r=(X_k\leq \alpha)$, where $X \in \real$ is a feature and $\alpha \in \real$ is a constant. We can now define the set of all instances in the dataset that satisfy $r$: $$S_r=\{d \in D | X_{d,k} \leq \alpha \}$$
Now it is possible to calculate MAD with respect to rule $r$, by simply replacing all appearances of $D$ with $S_r$, in \ref{mad}. 

\subsection{Prime implicants}

The procedure of computing the prime implicants of an instance is a simple modification of the one we developed for computing counterfactuals. Let us assume the instance of interest is $\textbf{X}=(X_1,X_2,\cdots,X_n)$. Furthermore, without loss of generality, we can assume that it is classified as 1, by the model. To compute the prime implicants of X, we have to form the objective function (with all coefficients equal to 1) and all the constraints that are necessary so the solution to the optimization problem is classified as 1, both of which should be performed in the same way as discussed in the main text. Finally, instead of minimizing this function, we have to maximize it. Intuitively, by doing so, we ask what is the largest set of features that can change values, without altering the model's decision. 

Let us assume that the solution to this problem dictates that variables $X_{i_1},X_{i_2},\cdots, X_{i_k}$, should change values. while the variables in $\textbf{Z}=\{X_1,X_2,\cdots,X_n\}\setminus \{X_{i_1},X_{i_2},\cdots, X_{i_k}\}$ should not. Then $\textbf{Z}$ is equal to the prime implicants. To see this, let us assume that $\textbf{Z}$ contains $m$ elements, and that the number of prime implicants of $\textbf{X}$ are $l<m$. Then, this means that as long as these $l$ variables maintain their values, the model will classify the datapoint as 1. In turn, this means that all the remaining $n-l$ variables can switch values, and that this would be a feasible solution to the optimization problem of the previous paragraph. Now this leads to a contradiction, since by assumption the solution alters the values of $n-m$ variables, meaning that the inequality $n-m\geq n-l\Rightarrow l\geq m$ should hold, which is not possible. A similar argument makes sure that conditional prime implicants can be generated by just incorporating the constraint that the conditioning variable maintains its value. This can be readily done, since it exactly corresponds to adding a diversity constraint.

\subsection{Additional case studies}
Tables 4, 5, 6 contain more case studies for the 3 datasets we consider. In table 6, the condition in all instances is that the vote on the first topic should be part of the resulting prime implicants.

\begin{table}[t]
\centering
\scalebox{0.75}{\begin{tabular}{|l|c|c|c|c|c|c||c|}
\hline
 & Sex & Age & Race & Juvenile felonies &  Prior crimes & Two year residivism & Outcome\\
\hline  
Factual & Male & 21 & Black & 0 & 0 & Yes & High score\\
Counterfactual  & Male & $\mathbf{\leq 19.5}$ & Black & 0 & 0 & Yes & Low score\\
Diverse counterfactual & Male & (\underline{>20}) \textbf{> 21.5} & Black &  0 & 0 & Yes & Low score\\
\hline  
\hline
Factual & Male & 27 & Black & 0 & 0 & No & Low score\\
Counterfactual  & Male & 27 & Black & 0 & \textbf{>4.5} & No & High score\\
Diverse counterfactual & Male & \textbf{<21.5} & Black & 0 & (\underline{=0}) 0 & No & High score\\
\hline
\hline 
Factual & Male & 32 & Black & 0 & 0 & No & Low score\\
Counterfactual  & Male & 27 & Black & 0 & \textbf{>1.5} & No & High score\\
Diverse counterfactual & Male & 32 & \textbf{Caucasian} & $\mathbf{\geq 1}$ & (\underline{$\leq 1$}) 0 & No & High score\\
\hline
\hline  
Factual & Male & 43 & Caucasian & 0 & 2 & No & Low score\\
Counterfactual  & \textbf{Female} & 43 & Caucasian & 0 & 2 & No & High score\\
Diverse counterfactual & (\underline{Male}) & 43 & Caucasian &  0 & \textbf{>4} & No & High score\\
\hline
\end{tabular}}
\caption{COMPAS dataset instances}
\label{}
\end{table}

\begin{table}[t]
\centering
\scalebox{0.75}{\begin{tabular}{|l|c|c|c|c|c|c||c|}
\hline
 & Sex & LSAT & Race & UGPA  & Outcome\\
\hline
Factual & Male & 36.5 & White & 3.2 & Pass\\  
Counterfactual  & Male & $\mathbf{20.75}$ & \textbf{Black} & $\mathbf{\leq 1.95}$ & Fail\\
Diverse counterfactual & Male & $\mathbf{19.25}$ & (\underline{White}) White & $\mathbf{2.15}$ &  Fail\\
\hline
\hline  
Factual & Female & 43 & White & 2.8 & Pass\\  
Counterfactual  & Female & $\mathbf{\leq 26.75}$ & White & 2.8 & Fail\\
Diverse counterfactual & Female & $\mathbf{\leq 19.25}$ & (\underline{White}) White & $\mathbf{\leq 2.15}$ &  Fail\\
\hline
\hline    
Factual & Male & 35 & White & 2.7 & Pass\\  
Counterfactual  & Male & 35 & \textbf{Black} & $\mathbf{\leq 1.95}$ & Fail\\
Diverse counterfactual & Male  & $\mathbf{\leq 19.25}$ & (\underline{White}) White & 2.7 &  Fail\\
\hline
\hline
Factual & Male & 33 & White & 3 & Pass\\   
Counterfactual  & Male & 33 & \textbf{Black} & $\mathbf{\leq 1.85}$ & Fail\\
Diverse counterfactual & Male  & $\mathbf{\leq 19.25}$ & (\underline{White}) White & 3 &  Fail\\
\hline
\end{tabular}}
\caption{LSAT dataset instances}
\label{}
\end{table}

\begin{table}[t]
\centering
\scalebox{0.75}{\begin{tabular}{|l|c|c|c|c|c|c|c|c|c|c|c|c|c|c|c|c||c|}
\hline
 & 1 & 2 & 3 & 4 & 5 & 6 & 7 & 8 & 9 & 10 & 11 & 12 & 13 & 14 & 15 & 16 & Outcome\\
\hline
Factual & + & + & - & + & + & + & - & - & - & + & - & + & - & + & - & + & Republican\\
Prime implicants  &  &  &  & \checkmark & \checkmark &  &  &  &  &  &  & \checkmark & \checkmark & \checkmark &  &  & Republican\\
Conditional prime implicants  & \checkmark &  & \checkmark & \checkmark & \checkmark & \checkmark &  & \checkmark &  &  &  &  &  & \checkmark &  &  & Republican\\
\hline
\hline
Factual & + & - & + & - & - & - & - & + & + & + & - & - & - & - & + & + & Democrat\\
Prime implicants  &  &  &  & \checkmark & \checkmark &  &  &  &  &  &  &  &  & \checkmark &  &  & Democrat\\
Conditional prime implicants  & \checkmark &  & \checkmark & \checkmark & \checkmark &  &  &  &  &  &  &  &  &  &  &  & Democrat\\
\hline
\hline
Factual & - & - & - & + & + & + & - & - & - & - & - & + & + & + & - & + & Republican\\
Prime implicants  &  &  &  & \checkmark & \checkmark &  &  &  & \checkmark &  &  &  & \checkmark & \checkmark & \checkmark &  & Republican\\
Conditional prime implicants  & \checkmark  &  & \checkmark & \checkmark & \checkmark &  &  &  &  &  &   &  & \checkmark & \checkmark &  &  & Republican\\
\hline
\hline
Factual & - & - & + & - & - & + & + & + & + & + & + & - & - & - & + & + & Democrat\\
Prime implicants  &  &  &  & \checkmark & \checkmark &  &  &  &  &  &  &  &  & \checkmark &  &  & Democrat\\
Conditional prime implicants  & \checkmark  &  & \checkmark & \checkmark & \checkmark &  &  &  &  &  &  &  &  & \checkmark &  &  & Democrat\\
\hline
\end{tabular}}
\caption{Congressional Voting Records dataset instances}
\label{}
\end{table}

\subsection{Proofs}

The proofs of propositions 2, 3 and 4 are immediate and follow directly form the definitions.

Proof of proposition 5:

\begin{proof}
Let $\textbf{X}$ be  an assignment that satisfies all the constraints. From the constraint $\sum_{i=1}^n \delta_i=1$ we have that there is a $j$, such that $\delta_j=1$. This $\delta_j$ appears in an additional constraint of the form $X_{j1} + X_{j2}\cdots + X_{jp}  \geq p\cdot\delta_j \Rightarrow X_{j1} + X_{j2}\cdots + X_{jp}  \geq p$. However, it also holds that $X_{j1} + X_{j2}\cdots + X_{jp}  \leq p$, so putting these two expressions together we have that $X_{j1} + X_{j2}\cdots + X_{jp}  = p \Rightarrow X_{j1} \cdot X_{j2}\cdots  X_{jp} = 1$, by proposition 4, which means that $P_0(\textbf{X})=1$. \qed
\end{proof}

Proof of proposition 6:

\begin{proof}
If $\delta=0$, the constraints can be rewritten as:

\begin{align*}
&X_{11} + X_{12}\cdots + X_{1k}  \leq k -1\\
&X_{21} + X_{22}\cdots + X_{2m} \leq m -1\\
&~~~~~~~~~~~~~~~~~~\cdots\\
&X_{n1} + X_{n2}\cdots + X_{nl}  \leq l -1
\end{align*}
By proposition 4, this means that all terms in the $0$-DP are zero, so $P_0(\textbf{X})=0$.\qed
\end{proof}

Proof of proposition 7:

\begin{proof}
The last constraint enforces that more that half of the $\delta_i$'s are equal to $1$. The result follows, since each $delta_i$ is an indicator a DT's outcome. This means that the majority of the DTs  classify the resulting instance in the desired category. \qed
\end{proof}

Proof of proposition 9:

\begin{proof}
Let $X\leq a$ be a feature in a RF. The first case we are going to examine is when this rule is not satisfied, meaning that $\mathbbm{1}[X_1\leq a]=0$. Then, the first constraint reduces to $\sum_{f_i \in F_x^+ (X\leq a)} \mathbbm{1}[f_i] \geq 0$, which always holds. The second constraint however becomes $\sum_{f_i \in F_x^- (X\leq a)} \mathbbm{1}[f_i] \leq 0$, which means that $\sum_{f_i \in F_x^- (X\leq a)} \mathbbm{1}[f_i] =0$, forcing all rules within $F_x^- (X\leq a)$ to be false as well, thus guaranteeing consistency wrt to the feature $X\leq a$.

The other case we need to examine is when $\mathbbm{1}[X_1\leq a]=1$. Then, the first constraint becomes $\sum_{f_i \in F_x^+ (X\leq a)} \mathbbm{1}[f_i] \geq |F_x^+ (X\leq a)|$, which implies that all rules within $F_x^+ (X\leq a)$ are also satisfied. The second constraint becomes $\sum_{f_i \in F_x^- (X\leq a)} \mathbbm{1}[f_i] \leq |F_x^- (X\leq a)|$, which always holds. \qed
\end{proof}

\end{document}